\documentclass[letterpaper]{article} 
\usepackage{aaai2027}  
\usepackage[hyphens]{url}  
\usepackage{graphicx} 
\usepackage{natbib}  
\usepackage{caption} 
\usepackage{algorithm}
\usepackage{algorithmic}

\usepackage{newfloat}
\usepackage{listings}
\DeclareCaptionStyle{ruled}{labelfont=normalfont,labelsep=colon,strut=off} 
\floatstyle{ruled}
\newfloat{listing}{tb}{lst}{}
\floatname{listing}{Listing}

\usepackage{booktabs}

\usepackage{booktabs} 
\usepackage{multirow}
\usepackage{tabularx}
\usepackage{makecell} 
\usepackage[table]{xcolor}
\usepackage{enumitem}

\usepackage{amsmath}  
\usepackage{amssymb} 
\usepackage{bm}      

\usepackage{footnote} 

\definecolor{mygray}{gray}{.9}

\nocopyright 

\title{Uncertainty-Aware 4D Gaussian Splatting for Monocular \\ Occluded Human Rendering}
\author{
    Weiquan Wang\textsuperscript{\rm 1}, 
    Feifei Shao\textsuperscript{\rm 1}, 
    Lin Li\textsuperscript{\rm 2}, 
    Zhen Wang\textsuperscript{\rm 2}, 
    Fengda Zhang\textsuperscript{\rm 3}, 
    Jun Xiao\textsuperscript{\rm 1}\corresponding, 
    Long Chen\textsuperscript{\rm 2}
}
\affiliations{
    \textsuperscript{\rm 1}Zhejiang University\\
    \textsuperscript{\rm 2}The Hong Kong University of Science and Technology\\
    \textsuperscript{\rm 3}Nanyang Technological University\\

    wqwangcs@zju.edu.cn, junx@cs.zju.edu.cn
}

\begin{document}

\maketitle

\begin{figure*}
  \centering
  \includegraphics[width=\textwidth]{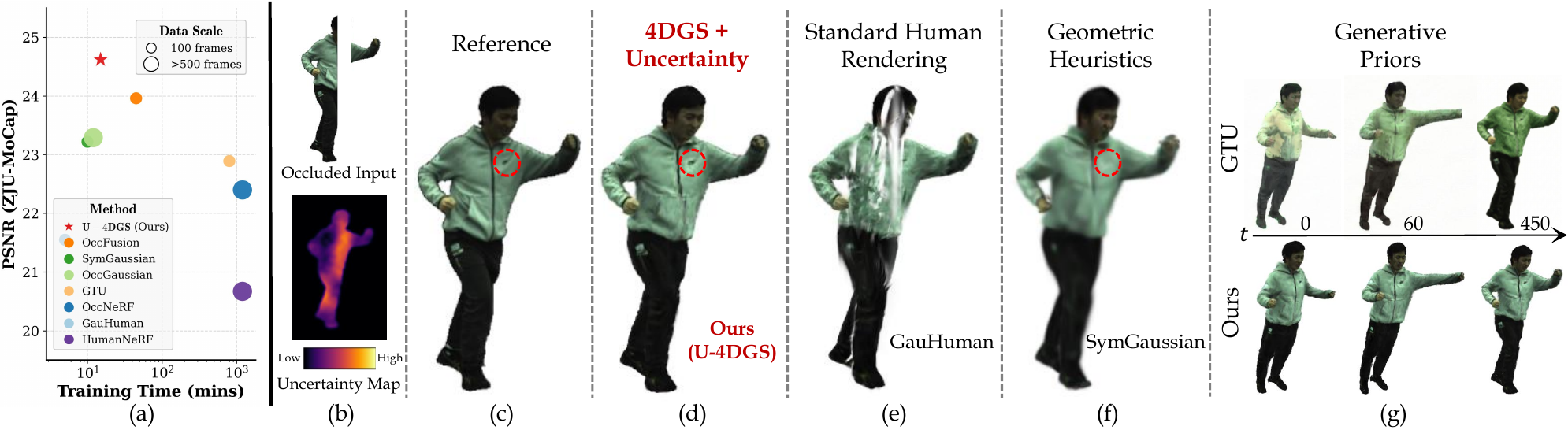}
  \caption{\textbf{Performance, Fidelity, and Stability.} 
  \textbf{(a)} Our U-4DGS achieves the best trade-off between rendering quality and training efficiency. 
  \textbf{(b)} Visualizes the occluded input and the corresponding uncertainty map predicted by our method.
  \textbf{(c)-(f)} Visual comparisons. While our method \textbf{(d)} recovers fine details consistent with the Reference \textbf{(c)}, 
  Gauhuman \textbf{(e)} fails catastrophically, fusing occlusion artifacts into the body.
  SymGaussian \textbf{(f)} fails to recover asymmetric details (see red circle), as the model erroneously propagates features from the unadorned opposite side.
  \textbf{(g)} Temporal consistency. GTU suffers from severe texture drifting, where the shirt color inconsistently shifts over time. In contrast, our method ensures physical consistency.}
  \label{fig:teaser}
\end{figure*}

\begin{abstract}
High-fidelity rendering of dynamic humans from monocular videos typically degrades catastrophically under occlusions. Existing solutions incorporate external priors—either hallucinating missing content via generative models, which induces severe temporal flickering, or imposing rigid geometric heuristics that fail to capture diverse appearances. To this end, we reformulate the task as a Maximum A Posteriori estimation problem under heteroscedastic observation noise. In this paper, we propose \textbf{U-4DGS}, a framework integrating a Probabilistic Deformation Network and a Joint Rasterization pipeline. This architecture renders pixel-aligned uncertainty maps that act as an adaptive gradient modulator, automatically attenuating artifacts from unreliable observations. Furthermore, to prevent geometric drift in regions lacking reliable visual cues, we enforce Confidence-Aware Regularizations, which leverage the learned uncertainty to selectively propagate spatial-temporal validity. Extensive experiments on the ZJU-MoCap and OcMotion datasets demonstrate that U-4DGS achieves SOTA rendering fidelity and robustness.
\end{abstract}


\section{1. Introduction}
\label{sec:intro}

High-fidelity reconstruction of dynamic humans from monocular videos stands as a foundational technology for immersive applications, spanning telepresence, metaverse interaction, sports analysis, and virtual reality~\cite{tian2023recovering, yang2025innovative}.
Recent years have witnessed remarkable progress in this domain, enabling the synthesis of photorealistic digital avatars from monocular inputs~\cite{weng2022humannerf, wen2024gomavatar}.
However, these successes hinge on the idealized assumption of full visibility and isolation from the background. 
This assumption rarely holds in unconstrained in-the-wild scenarios, where subjects inevitably interact with their surroundings, such as sitting on chairs, navigating behind obstacles, or being partially obscured by environmental objects.
When standard pipelines are applied to occluded scenarios, their performance degrades catastrophically~\cite{xiang2023rendering, sun2024occfusion}.
As illustrated in Fig.~\ref{fig:teaser}(e), the model erroneously interprets the occluder as part of the human geometry, manifesting as incomplete geometry or severe texture artifacts.
Thus, enabling robust human rendering from occluded observations is imperative to bridge the gap between laboratory prototypes and real-world applications.

To address these specific challenges, pioneering NeRF-based approaches, such as OccNeRF~\cite{xiang2023rendering} and Wild2Avatar~\cite{xiang2025rendering}, explore human rendering from occluded monocular videos. 
By incorporating geometry priors or decoupling scene components, these methods demonstrate the feasibility of recovering occluded surfaces. 
However, despite their visual fidelity, they inherit the intrinsic computational bottlenecks of implicit neural representations. 
Specifically, the reliance on expensive volumetric rendering incurs prohibitive training and inference costs, precluding their deployment in real-time applications.

Recently, the advent of 3D Gaussian Splatting (3DGS)~\cite{kerbl20233d} shifted the paradigm, offering real-time rendering capabilities~\cite{fei20243d, feng2025flashgs}. 
To adapt 3DGS for occluded human rendering, concurrent works incorporate external priors to compensate for missing information. 
Methods relying on geometric heuristics enforce rigid rules, such as left-right symmetry~\cite{jiang2025symgaussian} or feature aggregation~\cite{ye2025occgaussian}. 
However, these hand-crafted spatial priors typically lack flexibility in modeling diverse human appearances. 
As shown in Fig.~\ref{fig:teaser}(f), the symmetry prior fails to recover the chest logo, erroneously copying unadorned features from the opposite side. 
Alternatively, methods leveraging generative priors utilize 2D diffusion models to hallucinate invisible regions~\cite{sun2024occfusion, lee2024guess}. 
Yet, the inherent stochasticity of frame-by-frame generation often induces severe temporal flickering. 
This is evident in Fig.~\ref{fig:teaser}(g), where the shirt color drifts inconsistently across frames, violating physical consistency. 
Crucially, these methods rely on imperfect external priors to infer missing content, neglecting the intrinsic reliability of the input data, where valid cues are embedded across the temporal sequence.

To bridge this gap, our insight stems from the video's intrinsic temporal redundancy: since occlusions are transient, parts obscured in one frame are likely visible in others. 
This motivates adopting a 4D representation to consolidate fragmented observations into a unified canonical space. 
However, naive 4D Gaussian Splatting faces a supervision dilemma~\cite{wu20244d, li2024st}. 
Indiscriminately treating all pixels as valid constraints forces the deformation field to establish erroneous correspondences, compelling the human geometry to adhere to the occluder's surface texture. 
Conversely, a seemingly straightforward alternative is to apply hard visibility masks to exclude occluded regions from the optimization. 
Yet, strictly limiting supervision to visible regions leaves the occluded parts mathematically unconstrained; without gradient feedback, the geometry in these blind spots fails to densify or maintain coherence, resulting in incomplete structures.
Thus, rather than relying on brittle hard masks, the reconstruction process demands a dynamic discrimination mechanism to assess the validity of the observation, differentiating between visible human details and misleading observational noise (i.e., occluded regions).

To this end, we reformulate the task as a \textbf{Maximum A Posteriori (MAP) estimation} problem under heteroscedastic observation noise. Specifically, instead of assuming constant variance, we model the likelihood of each pixel as a Laplacian distribution with a learnable scale. This formulation naturally derives a probabilistic objective where the predicted scale—representing the aleatoric uncertainty—emerges as an inverse weighting term in the loss function. 
Consequently, this uncertainty functions as an \emph{adaptive gradient modulator}: low uncertainty permits RGB supervision to dominate the optimization for fine-grained refinement, while high uncertainty automatically attenuates the gradients from unreliable observations. 
As visualized in Fig.~\ref{fig:teaser}(b), the network explicitly captures this reliability distribution, autonomously assigning high uncertainty (bright regions) to 
occluded areas.

Building upon this formulation, we present \textbf{U-4DGS}, a robust uncertainty-aware framework for monocular occluded human rendering. To parameterize the heteroscedastic noise model, we propose a \emph{\textbf{Probabilistic Deformation Network}}. Beyond standard geometric warping, this module explicitly predicts the per-primitive aleatoric uncertainty, serving as the confidence metric for our MAP objective. 
To enable the adaptive gradient modulation, we introduce a \emph{\textbf{Joint Rasterization}} pipeline. By rendering pixel-aligned uncertainty maps alongside color, this mechanism enforces the inverse weighting scheme, ensuring that gradients from occluded regions are effectively attenuated. 
Finally, to prevent geometric drift in areas lacking visual cues (i.e., occluded regions), we design \emph{\textbf{Confidence-Aware Regularizations}}. These constraints leverage the learned uncertainty to selectively propagate validity from confident regions to unreliable areas, ensuring coherent completion. As illustrated in Fig.~\ref{fig:teaser}(a), U-4DGS achieves high-fidelity rendering even under severe occlusion while maintaining training efficiency.

In summary, our main contributions are as follows:
\begin{itemize}[leftmargin=*]
\item We reformulate occluded human rendering as a MAP estimation problem. By modeling heteroscedastic observation noise, we equip our optimization to discriminate between valid supervision and occlusion artifacts.

\item We propose U-4DGS, a framework integrating a Probabilistic Deformation Network with a Joint Rasterization pipeline. This design renders pixel-aligned uncertainty maps that act as adaptive gradient modulators, effectively shielding the canonical geometry from corruption.

\item We introduce Confidence-Aware Regularizations to prevent geometric drift in regions lacking reliable visual cues. Leveraging learned uncertainty, these constraints enforce spatial-temporal consistency for coherent completion.
    
\item We demonstrate that U-4DGS achieves state-of-the-art rendering fidelity on ZJU-Mocap and OcMotion datasets, successfully recovering clean avatars from severe occlusions.
\end{itemize}

\section{2. Related Works}

\noindent\textbf{3D Human Avatar Reconstruction.}
Traditional human reconstruction methods relying on dense camera arrays or depth sensors struggle in in-the-wild scenarios~\cite{collet2015high, su2020robustfusion, huang2020arch, yu2021function4d}. 
To address this, NeRF-based methods have enabled high-fidelity reconstruction from monocular inputs by conditioning radiance fields on SMPL priors~\cite{yu2023monohuman, te2022neural, li2024ghunerf}. 
However, the reliance on volumetric ray-marching incurs prohibitive computational costs, precluding real-time applicability. 
Recently, 3D Gaussian Splatting (3DGS)~\cite{kerbl20233d} has introduced a paradigm shift with its real-time rasterization capabilities. Building upon the skeleton-conditioned canonical deformation paradigm pioneered by early implicit frameworks like Animatable NeRF~\cite{peng2021animatable}, SOTA methods~\cite{qian20243dgs, kocabas2024hugs, hu2024gauhuman, lei2024gart, shao2024splattingavatar, zhao2025surfel} adapt this representation to dynamic humans by representing the body in a canonical space and transforming it via Linear Blend Skinning. 
Critically, while performant under full visibility, these pipelines implicitly assume valid supervision for all pixels. Lacking mechanisms to handle occlusions, they catastrophically fuse occluder artifacts into the avatar geometry.

\noindent\textbf{Occlusion-Aware Human Rendering.}
Existing methods generally fall into three paradigms. 
Early strategies focused on \emph{scene decoupling}, employing visibility priors to isolate the subject from obstacles~\cite{xiang2025rendering, xiang2023rendering}; however, these implicit frameworks incur high computational overhead. To synthesize unobserved content, recent approaches leverage \emph{generative priors} via pre-trained 2D diffusion models~\cite{sun2024occfusion, lee2024guess, fan2026inpainthuman}. Yet, the inherent stochasticity of such processes frequently induces identity drift and temporal flickering. 
Alternatively, other 3DGS-based methods employ \emph{geometric heuristics} such as symmetry constraints~\cite{ye2025occgaussian, jiang2025symgaussian}, which often falter under asymmetric clothing or motions. Distinctively, U-4DGS eschews unstable hallucinations and rigid heuristics.
By formulating reconstruction as a MAP estimation and aggregating valid temporal information via adaptive gradient modulation, we ensure physically consistent and subject-faithful results.

\noindent\textbf{Dynamic Reconstruction and Uncertainty Modeling.}
Recent strides in dynamic Gaussian Splatting have achieved high-fidelity 4D reconstruction via deformation fields~\cite{yang2024deformable, wu20244d, luiten2024dynamic, bae2024per, zhang2025mega}.
However, relying on deterministic canonical-observation correspondences causes solvers to overfit to occluders when valid motion cues are absent~\cite{guo2025uncertainty}.
To mitigate this, Bayesian aleatoric uncertainty~\cite{kendall2017uncertainties} is widely explored to enhance robustness against transient noise~\cite{sunderhauf2023density, goli2024bayes, ren2024nerf}, utilizing strategies like per-image embeddings~\cite{martin2021nerf}, robust losses~\cite{sabour2023robustnerf}, or density parameterizations~\cite{lee2025bayesian}.
Concurrently, USPLAT4D~\cite{guo2025uncertainty} employs uncertainty-aware motion graphs for general scenes. Yet, treating uncertainty merely as a smoothing weight is insufficient for articulated humans, where structural integrity is paramount. 
In contrast, U-4DGS elevates uncertainty to actively modulate optimization: it shields canonical geometry from misleading gradients and triggers confidence-aware regularizations to enforce physical plausibility in occluded regions.

\begin{figure*}[t]
    \centering
    \includegraphics[width=1.0\linewidth]{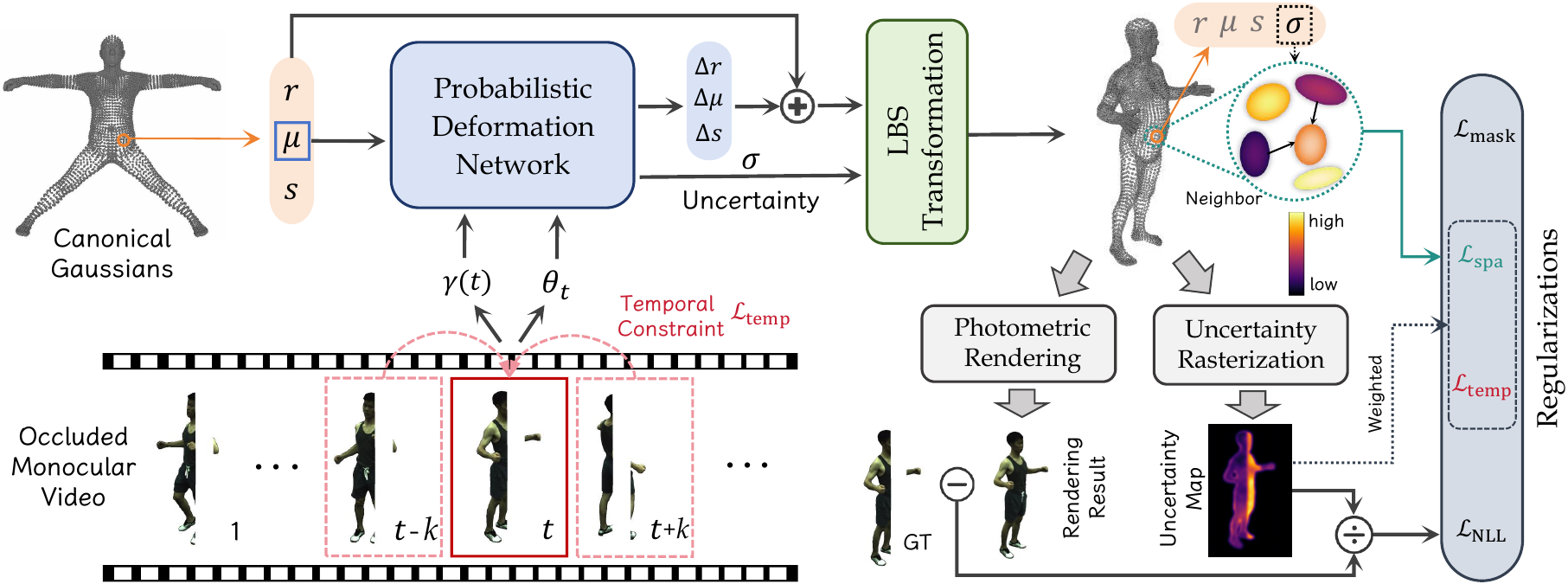}
    \caption{\textbf{The framework of U-4DGS.} 
    \textbf{Left:} The \emph{Probabilistic Deformation Network} conditions Canonical Gaussians on time embedding $\gamma(t)$ and pose $\theta_t$ to predict geometric offsets ($\Delta \mathbf{r}, \Delta \mathbf{\mu}, \Delta \mathbf{s}$) alongside per-primitive aleatoric uncertainty $\sigma$. 
    \textbf{Middle:} The deformed Gaussians are transformed via LBS and rendered through a \emph{Joint Rasterization} pipeline, simultaneously producing a photometric image and a pixel-aligned Uncertainty Map (where bright regions indicate high uncertainty).
    \textbf{Right:} During optimization, the Uncertainty Map functions as an adaptive gradient modulator (symbolized by $\div$) in the $\mathcal{L}_{NLL}$ objective, effectively attenuating gradients from unreliable observations. Simultaneously, \emph{Confidence-Aware Regularizations} ($\mathcal{L}_{spa}, \mathcal{L}_{temp}$) leverage the learned uncertainty to enforce spatial-temporal constraints in regions lacking reliable visual cues.}
    \label{fig:pipeline}
\end{figure*}

\section{3. Preliminaries}
\label{sec:preliminaries}

\noindent\textbf{Articulated 3D Gaussian Splatting.} Following recent practices~\cite{hu2024gauhuman, qian20243dgs}, we model dynamic humans by instantiating 3D Gaussians within a canonical SMPL~\cite{loper2015smpl} space, binding each primitive $i$ to its nearest mesh vertex. 
Each Gaussian is parameterized by its canonical mean $\mathbf{\mu}_i \in \mathbb{R}^3$, scaling $\mathbf{s}_i \in \mathbb{R}^3$, rotation quaternion $\mathbf{q}_i \in \mathbb{R}^4$, opacity $\alpha_i \in [0, 1]$, and color $\mathbf{c}_i$. 
For a frame $t$ with SMPL pose $\theta_t$ and shape $\beta$, the canonical geometry is transformed to the observation space via Linear Blend Skinning (LBS): $\mathbf{x}_{obs} = W(\mathbf{x}_{can}, \mathbf{J}(\beta), \theta_t, \mathcal{W})$. 
The deformed primitives are then projected onto the image plane to render the pixel color via differentiable $\alpha$-blending: 
\begin{equation} 
    \mathbf{C}(\mathbf{u}) = \sum_{i \in \mathcal{N}} T_i \alpha_i G_i^{2D}(\mathbf{u}) \mathbf{c}_i, 
\label{eq:splatting} 
\end{equation}
where $G_i^{2D}(\mathbf{u})$ is the 2D Gaussian density and $T_i = \prod_{j=1}^{i-1} (1 - \alpha_j G_j^{2D}(\mathbf{u}))$ denotes the accumulated transmittance.

\noindent\textbf{Problem Formulation.} Given a monocular video $\mathcal{V} = \{\mathbf{I}_t\}_{t=1}^T$ with coarse SMPL tracking, standard 3DGS optimization can be viewed as a Maximum Likelihood Estimation (MLE) problem. 
Minimizing the photometric loss implicitly assumes that observation noise across all pixels follows a fixed, homoscedastic (constant variance) distribution. 
However, under environmental occlusion, observation noise is inherently heteroscedastic: visible regions exhibit low noise, whereas occluded regions introduce gross corruption. 
Applying a homoscedastic MLE objective forces the model to overfit these outliers, erroneously adhering the human geometry to occluders.


To resolve this, we reformulate the reconstruction as a \textbf{Maximum A Posteriori (MAP)} estimation. 
Our goal is to jointly estimate the canonical parameters $\Theta$ and the per-pixel aleatoric uncertainty (representing the noise scale). 
This formulation allows us to dynamically down-weight unreliable observations (the Likelihood term) while incorporating structural priors to constrain regions lacking valid visual cues (the Prior term), as detailed in Section Methodology.


\section{4. Methodology}
\label{sec:method}

\noindent\textbf{Overview.} Building upon the probabilistic formulation, we propose \textbf{U-4DGS}, a framework designed to realize the MAP estimation for robust occluded human rendering. 
U-4DGS explicitly models the heteroscedastic nature of observation noise, enabling an autonomous arbitration mechanism between valid supervision and occlusion artifacts.
As illustrated in Fig.~\ref{fig:pipeline}, the pipeline consists of three integral components:
\emph{\textbf{(1) Probabilistic Deformation:}} We introduce a network $\Phi$ that serves a dual role: it models the temporal geometry evolution via deformation offsets and parameterizes the input-dependent aleatoric uncertainty $\sigma$ for each Gaussian, explicitly quantifying the confidence of the temporal correspondence.
\emph{\textbf{(2) Joint Rasterization:}} To propagate this 3D confidence to the 2D observation space, we employ a dual-branch rasterizer. This module renders the deformed geometry into a photometric image $\hat{\mathbf{C}}$ while simultaneously accumulating per-primitive uncertainty into a pixel-aligned Uncertainty Map $\hat{U}$.
\emph{\textbf{(3) Optimization via MAP Estimation:}} These outputs drive our MAP-based objective. Here, $\hat{U}$ functions as an adaptive gradient modulator within the probabilistic photometric loss, attenuating gradients from unreliable observations. Simultaneously, to prevent geometric drift in regions lacking reliable visual cues, we enforce Confidence-Aware Regularizations that leverage the learned uncertainty to impose spatial-temporal priors.



\subsection{4.1 Probabilistic Deformation Network}

To instantiate the MAP framework, we require a mechanism that not only recovers time-varying geometry but also quantifies the reliability of each correspondence. Unlike standard deterministic fields~\cite{yang2024deformable, hu2024gauhuman} that indiscriminately fit all observations, we propose a Probabilistic Deformation Network $\Phi_\psi$. This module serves a dual purpose: (1) it models high-frequency non-rigid residuals beyond the SMPL prior; (2) it explicitly predicts the aleatoric uncertainty $\sigma$, providing the input-dependent noise scale required for our probabilistic objective.


We parameterize $\Phi_\psi$ as a Multi-Layer Perceptron (MLP). To capture high-frequency details in both spatial and temporal domains, we apply sinusoidal positional encoding $\gamma(\cdot) = \left( \sin(2^k \pi \cdot), \cos(2^k \pi \cdot) \right)^{L-1}_{k=0}$ to both the input coordinates and time. The network conditions the canonical Gaussians on these embeddings and the 3D skeleton pose $\theta_t$ to predict geometry offsets and uncertainty:
\begin{equation}
    (\Delta \mathbf{\mu}, \Delta \mathbf{r}, \Delta \mathbf{s}, \sigma) = \Phi_\psi(\gamma(\text{sg}(\mathbf{\mu}_{can})), \gamma(t), \theta_t),
\end{equation}
where $\text{sg}(\cdot)$ denotes the stop-gradient operator.
The offsets $(\Delta \mathbf{\mu}, \Delta \mathbf{r}, \Delta \mathbf{s})$ explicitly model the time-varying geometry, enabling the fusion of features across frames.
For the uncertainty $\sigma$, we employ a Softplus activation to ensure strictly positive values ($\sigma \in \mathbb{R}^+$).
We then apply the predicted offsets to the canonical Gaussians to obtain the deformed state:
\begin{equation}
    \mathbf{\mu}_{def} = \mathbf{\mu}_{can} + \Delta \mathbf{\mu}, \quad \mathbf{s}_{def} = \mathbf{s}_{can} + \Delta \mathbf{s}, \quad \mathbf{q}_{def} = \mathbf{q}_{can} \cdot \Delta \mathbf{r}.
\end{equation}
These deformed primitives are subsequently transformed to the observation space via LBS.

Crucially, since $\sigma$ is predicted in the canonical space, it is defined as an intrinsic property of the Gaussian primitive.
It remains bound to the surface during the LBS transformation, invariant to the global body movement.
This design allows the network to learn specific spatial-temporal failure modes, such as assigning consistently high uncertainty to body parts undergoing rapid deformation or frequent self-occlusion, regardless of their global position in the camera view.



\subsection{4.2 Joint Rasterization}
\label{sec:rasterization}

To bridge the gap between the 3D aleatoric uncertainty $\sigma$ and the 2D observation space required for our MAP objective, we propose a Joint Rasterization scheme. Since occlusion is an inherently view-dependent phenomenon, the reliability of a pixel is determined by the accumulated uncertainty along its corresponding optical ray. Therefore, we perform two pixel-aligned rasterization passes in parallel, generating both the photometric appearance $\hat{\mathbf{C}}$ and the uncertainty map $\hat{U}$.


The first pass renders the predicted color image $\hat{\mathbf{C}}$ using the standard splatting formulation (Eq.~\ref{eq:splatting}). Crucially, this pass utilizes the deformed geometry derived from the Probabilistic Deformation Network, ensuring that the rendering faithfully reflects the current temporal dynamics.


The second pass renders the uncertainty map $\hat{U} \in \mathbb{R}^{H \times W}$. To ensure geometric consistency between the rendered appearance and its associated confidence, we accumulate the uncertainty using the same $\alpha$-blending weights as color:
\begin{equation}
\hat{U}(\mathbf{u}) = \sum_{i \in \mathcal{N}} T_i \alpha_i \sigma_i,
\end{equation}
where $T_i$ and $\alpha_i$ are identical to those computed in the color pass.
This formulation has a clear physical interpretation: $\hat{U}(\mathbf{u})$ represents the expected aleatoric uncertainty of the visible surface at pixel $\mathbf{u}$. By compositing $\sigma$ with the opacity $\alpha$, the resulting map accurately reflects the confidence of the foremost visible surface, automatically disregarding occluded primitives hidden behind.
Consequently, $\hat{U}$ serves as the spatially varying noise scale, functioning as the adaptive gradient modulator in the subsequent optimization.


\begin{table*}[t]
    \centering
    \caption{\textbf{Quantitative comparison on ZJU-MoCap and OcMotion datasets.} LPIPS values are scaled by $\times 1000$. The best results are \textbf{bolded} and the second-best results are \underline{underlined}. * Metrics calculated on visible pixels only.}
    \label{tab:quantitative}
    
    \renewcommand{\arraystretch}{1.15} 
    \begin{tabular}{|r|c||ccc|ccc|}
    
    \noalign{\hrule height 1pt}
    
    \rowcolor{mygray}
     & & \multicolumn{3}{c|}{\textbf{ZJU-MoCap}} & \multicolumn{3}{c|}{\textbf{OcMotion}} \\
    
    \cline{3-8} 
    \rowcolor{mygray} 

    \multicolumn{1}{|c|}{\multirow{-2}{*}{\textbf{Method}}} & 
    \multicolumn{1}{c||}{\multirow{-2}{*}{\textbf{Category}}} & 
    PSNR $\uparrow$ & SSIM $\uparrow$ & LPIPS $\downarrow$ & PSNR* $\uparrow$ & SSIM* $\uparrow$ & LPIPS* $\downarrow$ \\
    \hline
    \hline
    
    \multicolumn{8}{|c|}{\textit{Standard Human Rendering}} \\
    \hline
    HumanNeRF \small{(CVPR 2022)} & - & 20.67 & 0.9509 & - & - & - & - \\
    GaussianAvatar \small{(CVPR 2024)} & - & 18.01 & 0.9512 & 60.33 & - & - & - \\
    GauHuman \small{(CVPR 2024)} & - & 21.55 & 0.9430 & 55.88 & 15.09 & 0.8525 & 107.1 \\
    \hline

    \multicolumn{8}{|c|}{\textit{Occlusion-Aware Approaches}} \\
    \hline
    
    OccNeRF \small{(ICCV 2023)} & 
    \multirow{2}{*}{\shortstack{Scene\\Decoupling}} & 
    22.40 & \underline{0.9562} & 43.01 & 15.71 & 0.8523 & 82.90 \\
    
    Wild2Avatar \small{(TPAMI 2025)} & 
     & 
    - & - & - & 14.09 & 0.8484 & 93.31 \\
    \hline
    
    OccGaussian \small{(ICME 2025)} & 
    \multirow{2}{*}{\shortstack{Geometric\\Heuristics}} & 
    23.29 & 0.9482 & 41.93 & - & - & - \\
    
    SymGaussian \small{(ICASSP 2025)} & 
     & 
    23.22 & 0.9535 & 39.02 & - & - & - \\
    \hline
    
    GTU \small{(CVPR 2024)} & 
    \multirow{2}{*}{\shortstack{Generative\\Priors}} & 
    22.89 & 0.9503 & 40.78 & 15.83 & 0.8437 & 83.46 \\
    
    OccFusion \small{(NeurIPS 2024)} & 
     & 
    \underline{23.96} & 0.9548 & \underline{32.34} & \underline{18.28} & \underline{0.8875} & \underline{82.42} \\
    
    \hline

    USplat4D \small{(ICLR 2026)} & 
    \multirow{2}{*}{\shortstack{Uncertainty\\-aware}} & 
    23.68 & 0.9465 & 38.24 & 17.86 & 0.8722 & 82.96 \\
    
    \textbf{U-4DGS (Ours)} & 
    & 
     \textbf{24.62} & \textbf{0.9606} & \textbf{31.72} & \textbf{20.11} & \textbf{0.9030} & \textbf{79.17} \\ 

    
    
    \noalign{\hrule height 1pt}
    \end{tabular}
\end{table*}

\begin{table}[h]
    \centering
    \caption{\textbf{Quantitative evaluation on the occluded regions of ZJU-MoCap.} 
    Occ-PSNR and Occ-SSIM denote metrics calculated strictly within the ground-truth occlusion masks.}
    \label{tab:supp_occ_eval}
    
    \renewcommand{\arraystretch}{1.1}
    \setlength{\tabcolsep}{10pt} 
    \begin{tabular}{|l||cc|}
    \noalign{\hrule height 1pt}
    \rowcolor{mygray}
    \multicolumn{1}{|c||}{\textbf{Method}} & \textbf{Occ-PSNR} $\uparrow$ & \textbf{Occ-SSIM} $\uparrow$ \\
    \hline
    \hline
    OccNeRF & 20.48 & 0.9453 \\
    OccGaussian & 22.15 & 0.9387 \\
    SymGaussian & 22.06 & 0.9436 \\
    GTU & 21.83 & 0.9421 \\
    OccFusion & 22.84 & 0.9459 \\
    USplat4D  & 21.65 & 0.9361 \\
    \hline
    \textbf{U-4DGS (Ours)} & \textbf{23.27} & \textbf{0.9477} \\
    \noalign{\hrule height 1pt}
    \end{tabular}
\end{table}

    
    
    
    
    

\begin{figure*}[t]
    \centering
    \includegraphics[width=0.90\linewidth]{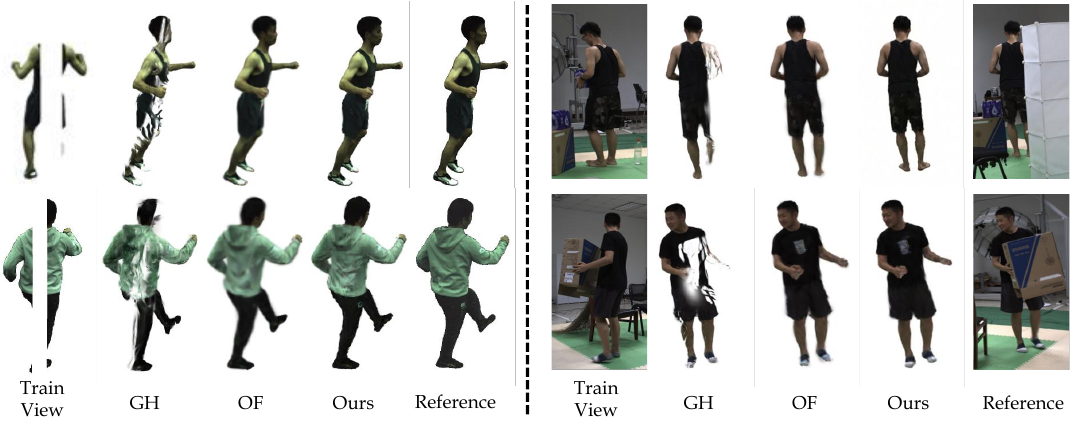}
    \caption{\textbf{Qualitative comparisons on novel view synthesis.} 
    \textbf{Left:} Results on the ZJU-MoCap dataset with synthetic occlusions. 
    \textbf{Right:} Results on the OcMotion dataset with real-world occlusions.
    \textbf{GH} denotes GauHuman~\cite{hu2024gauhuman} and \textbf{OF} denotes OccFusion~\cite{sun2024occfusion}. More qualitative results are shown in the supplementary material.
    }
    \label{fig:qualitative}
\end{figure*}

\subsection{4.3 Optimization via MAP Estimation}
\label{sec:optimization}

To reconstruct high-fidelity humans from monocular videos, we jointly optimize the Gaussian parameters $\Theta$ and the deformation network weights $\psi$. Following the MAP formulation introduced in Sec.~\ref{sec:preliminaries}, our objective function maximizes the posterior $P(\Theta|\mathcal{D}) \propto P(\mathcal{D}|\Theta)P(\Theta)$.
This decomposes the optimization into two synergistic components: a Likelihood term $P(\mathcal{D}|\Theta)$ that models data fidelity under heteroscedastic noise, and a Prior term $P(\Theta)$ that enforces physical constraints via Confidence-Aware Regularizations.


\noindent\textbf{Uncertainty-Weighted Photometric Loss ($P(\mathcal{D}|\Theta)$).}
The primary challenge in occluded rendering is the mismatch between the reconstructed geometry and occluder-corrupted observations. Standard objectives (e.g., L1 loss) assume homoscedastic noise, treating all pixels equally and thus forcing the model to fit occlusion artifacts.
To resolve this, we model the observation likelihood $P(\mathcal{D}|\Theta)$ by assuming the pixel-wise residual follows a \textit{Laplacian distribution}, where the location parameter is the predicted color $\hat{\mathbf{C}}(\mathbf{u})$ and the scale parameter is the predicted uncertainty $\hat{U}(\mathbf{u})$. The probability density function is given by $p(x|\mu) = \frac{1}{2\hat{U}}\exp(-\frac{|x-\mu|}{\hat{U}})$.
Minimizing the negative log-likelihood (NLL) of this distribution yields our uncertainty-weighted objective:
\begin{equation}
\small
    \mathcal{L}_{NLL} = \sum_{\mathbf{u} \in \Omega} \left( \frac{\| \mathbf{C}_{gt}(\mathbf{u}) - \hat{\mathbf{C}}(\mathbf{u}) \|_1}{\hat{U}(\mathbf{u}) + \epsilon} + \lambda_{reg} \log (\hat{U}(\mathbf{u}) + \epsilon) \right),
    \label{eq:nll}
\end{equation}
where $\epsilon=10^{-7}$ ensures numerical stability.
This objective acts as an adaptive gradient modulator: in visible regions, the solver minimizes the numerator ($\mathcal{L}_1$ error) and reduces $\hat{U}$; in occluded regions, it autonomously increases $\hat{U}$ to attenuate the gradient magnitude. This mechanism effectively shields the canonical geometry from misleading supervision. 

\noindent\textbf{Confidence-Aware Regularizations ($P(\Theta)$).}
While the uncertainty-weighted likelihood effectively attenuates gradients from artifacts, it leaves regions lacking reliable visual cues mathematically unconstrained. Without valid supervision, the human Gaussians in these blind spots are prone to the ill-posed geometric drift mentioned in Sec.~\ref{sec:intro}.
To resolve this, we impose the prior $P(\Theta)$ via Confidence-Aware Regularizations. These constraints utilize the learned uncertainty $\sigma$ to selectively enforce physical plausibility: applying strong regularization only where observations are unreliable, while allowing data-driven deformation in confident regions.


\textit{(1) Confidence-Guided Spatial Consistency:}
To maintain geometric integrity, we enforce local rigidity weighted by uncertainty. We construct a KNN graph in the canonical space. For each Gaussian $i$ and its neighbor $j \in \mathcal{N}_{can}(i)$, we penalize the discrepancy of their deformation attributes $\mathcal{F} = \{ \Delta \mathbf{\mu}, \Delta \mathbf{r}, \Delta \mathbf{s} \}$:
\begin{equation}
    \mathcal{L}_{spa} = \sum_{i} \text{sg}(\sigma_i) \sum_{j} \omega_{ij} \sum_{v \in \mathcal{F}} \lambda_v \| v_i - v_j \|_2,
\label{eq:spa}
\end{equation}
where $\omega_{ij}$ weights neighbors by canonical distance, and $\lambda_v$ balances the respective deformation components.
The term $\text{sg}(\sigma_i)$ acts as an attention mechanism: When a Gaussian is occluded ($\sigma_i$ is high), the $\mathcal{L}_{NLL}$ gradients vanish, and $\mathcal{L}_{spa}$ penalty dominates, forcing the Gaussian to synchronize with its neighbors.
Crucially, since visible neighbors are firmly anchored by the photometric loss, this constraint effectively propagates geometric validity from the confident boundary inward to the occluded regions.



\textit{(2) Uncertainty-Weighted Temporal Inertia:}
To prevent jittering in the absence of visual cues, we enforce a smoothness prior on the trajectory of uncertain Gaussians. 
For video sequences with a frame interval $k$, we formulate this as a second-order difference over sampled valid timestamps $t \in \mathcal{T}'$:
\begin{equation}
    \mathcal{L}_{temp} = \frac{1}{|\mathcal{T}'|} \sum_{t \in \mathcal{T}'} \sum_{i} \text{sg}(\sigma_{i,t}) \| \mathcal{F}_{i,t-k} - 2\mathcal{F}_{i,t} + \mathcal{F}_{i,t+k} \|_2.
\label{eq:temp}
\end{equation}
This term acts as an inertial prior. For high-uncertainty Gaussians, it penalizes acceleration, effectively interpolating their dynamics based on the trajectory established by confident frames, ensuring smooth motion transition across occlusion gaps.

In both $\mathcal{L}_{spa}$ and $\mathcal{L}_{temp}$, we apply a stop-gradient operator $\text{sg}(\cdot)$ to the uncertainty weights $\sigma$ to prevent degenerate optimization. Because $\sigma$ scales the regularization penalties, unconstrained backpropagation would encourage the network to trivially drive $\sigma \to 0$ simply to minimize these losses, invalidating its role in modeling heteroscedastic observation noise. By decoupling the gradient flow, $\text{sg}(\cdot)$ ensures that the aleatoric uncertainty is learned exclusively from the photometric likelihood, functioning purely as a non-differentiable attention mask for the spatial-temporal constraints.

\noindent\textbf{Total Loss.} 
The final objective function combines the uncertainty-weighted likelihood (data fidelity), the confidence-aware priors (regularization), and auxiliary perceptual constraints:
\begin{equation}
    \mathcal{L}_{total} = \mathcal{L}_{nll} + \lambda_{spa} \mathcal{L}_{spa} + \lambda_{temp} \mathcal{L}_{temp} + \lambda_{mask} \mathcal{L}_{mask} + \mathcal{L}_{img},
\end{equation}
where $\mathcal{L}_{img} = \lambda_{\text{ssim}} \mathcal{L}_{\text{ssim}} + \lambda_{\text{lpips}} \mathcal{L}_{\text{lpips}}$ ensures perceptual quality using standard metrics~\cite{wang2004image, zhang2018unreasonable}.
Additionally, $\mathcal{L}_{mask} = \| \hat{\mathbf{O}} - \mathbf{M}_{smpl} \|_2$ imposes a global silhouette constraint, aligning the accumulated opacity  $\hat{\mathbf{O}}$ with the projected SMPL mask $\mathbf{M}_{smpl}$ to guide coarse geometry. $\lambda$ balances the contribution of each term.

\section{5. Experiments}
\label{sec:experiments}

\subsection{5.1 Experimental Setup}

\noindent\textbf{Datasets.} 
We evaluate U-4DGS on synthetic \textbf{\emph{ZJU-MoCap}} \cite{peng2021neural} and real-world \textbf{\emph{OcMotion}}~\cite{huang2022occluded} datasets. To ensure strict and fair comparisons, we directly adopt the established training and evaluation protocols from prior works: we follow OccNeRF~\cite{xiang2023rendering} for ZJU-MoCap and OccFusion~\cite{sun2024occfusion} for OcMotion.

\vspace{1mm}
\noindent\textbf{Baselines.} 
We benchmark against two categories of state-of-the-art methods: (1) \textbf{\emph{Standard Human Rendering}} without occlusion handling (HumanNeRF~\cite{weng2022humannerf}, GaussianAvatar~\cite{hu2024gaussianavatar}, GauHuman~\cite{hu2024gauhuman}); and (2) \textbf{\emph{Occlusion-Aware Approaches}}, encompassing \textit{scene decoupling} (OccNeRF~\cite{xiang2023rendering}, Wild2Avatar~\cite{xiang2025rendering}), \textit{geometric heuristics} (OccGaussian~\cite{ye2025occgaussian}, SymGaussian~\cite{jiang2025symgaussian}), \textit{generative priors} (OccFusion~\cite{sun2024occfusion}, GTU~\cite{lee2024guess}), and \textit{uncertainty-aware method} (USplat4D~\cite{guo2025uncertainty}). For fair comparison, all methods utilize identical segmentation masks and SMPL priors. Detailed descriptions of baselines are provided in the supplementary material.

\vspace{1mm}
\noindent\textbf{Implementation Details.}
To preclude trivial solutions where the model predicts infinite uncertainty to bypass the NLL loss, we adopt a progressive training strategy: the geometry is supervised solely via standard $L_1$ loss for the initial 2k iterations before activating the uncertainty head and switching to our MAP objective (Eq.~\ref{eq:nll}). Comprehensive details regarding networks, training schedules, and hyperparameters are extensively discussed in the supplementary material.

\begin{figure}[t]
    \centering
    \includegraphics[width=1.0\linewidth]{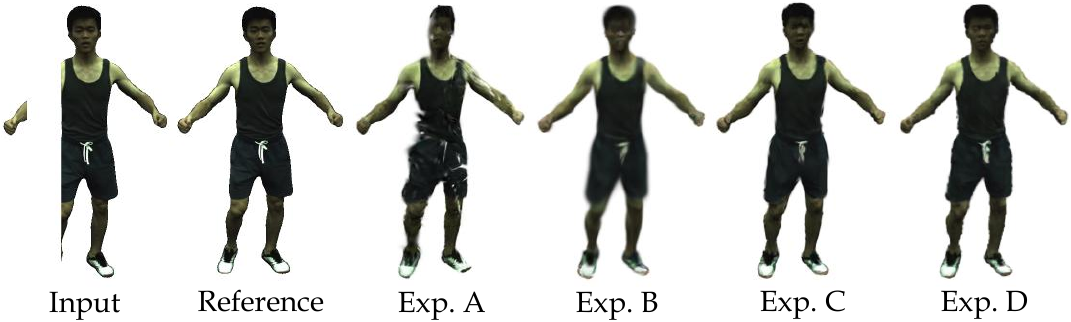} 
    \caption{\textbf{Qualitative ablation study.} 
    }    
    \label{fig:ablation}
\end{figure}

\begin{table}[t]
    \centering
    \caption{\textbf{Ablation study on the ZJU-Mocap dataset.} 
    $\mathcal{L}_{spa}$: Confidence-guided spatial constraint; 
    $\mathcal{L}_{temp}$: Uncertainty-weighted temporal inertia. constraint}
    \label{tab:ablation}
    \renewcommand{\arraystretch}{1.1}

    \begin{tabular}{|c||l|cc|}
    \noalign{\hrule height 1pt}
    \rowcolor{mygray}
    \textbf{Exp.} & \multicolumn{1}{c|}{\textbf{Configuration}} & \textbf{PSNR} $\uparrow$ & \textbf{SSIM} $\uparrow$ \\
    \hline
    \hline
    A & Baseline & 22.85 & 0.9490 \\
    B & + Uncert. Modeling & 24.15 & 0.9575 \\
    C & + Spa. Con. ($\mathcal{L}_{spa}$) & 24.48 & 0.9592  \\
    \hline
    D & + Temp. Con. ($\mathcal{L}_{temp}$) & \textbf{24.62} & \textbf{0.9606} \\
    \noalign{\hrule height 1pt}
    \end{tabular}
\end{table}

\subsection{5.2 Comparisons with State-of-the-Arts}
\label{sec:comparisons}

\noindent\textbf{Quantitative Results.} Tab.~\ref{tab:quantitative} reports performance on ZJU-MoCap and OcMotion. On the synthetic ZJU-MoCap, U-4DGS sets a new SOTA (24.62 dB PSNR), surpassing the leading generative baseline OccFusion (23.96 dB). This validates that uncertainty-guided aggregation yields superior fidelity. 
For the real-world OcMotion dataset (evaluated on visible pixels due to lack of occluded ground-truth), U-4DGS outperforms the runner-up by a substantial 1.83 dB. 

Crucially, to directly assess recovery capability, we compute metrics strictly within the occluded regions of ZJU-MoCap (Occ-PSNR). As shown in Tab.~\ref{tab:supp_occ_eval}, U-4DGS achieves 23.27 dB, surpassing both OccFusion and USplat4D.

\vspace{1mm}
\noindent\textbf{Qualitative Results.} 
Fig.~\ref{fig:qualitative} visualizes rendering quality. Standard methods like GauHuman (GH)~\cite{hu2024gauhuman} lack occlusion awareness, erroneously fusing obstacles (e.g., bars) onto the human body. Generative approaches like OccFusion (OF)~\cite{sun2024occfusion} remove occluders but suffer from stochasticity, yielding over-smoothed or identity-inconsistent hallucinations. Conversely, U-4DGS achieves superior fidelity. Our adaptive gradient modulator attenuates artifact gradients, while temporal aggregation faithfully restores clean, sharp, and physically consistent avatars even under severe occlusion. 

Furthermore, U-4DGS maintains rendering (155 FPS) with minimal training overhead (7.5 mins); detailed efficiency analysis is provided in the supplementary material.

\subsection{5.3 Ablation and Analysis}
\label{sec:ablation}

\noindent\textbf{Ablation Studies.} We incrementally validate each component on ZJU-MoCap (Tab.~\ref{tab:ablation}, Fig.~\ref{fig:ablation}).
\emph{\textbf{(1) Exp. A (Baseline):}} A deterministic baseline (adapted from GauHuman~\cite{hu2024gauhuman}) optimized via $L_1$ loss. Lacking occlusion awareness, it indiscriminately fits environmental obstacles, causing severe artifact adhesion.
\emph{\textbf{(2) Exp. B (+ Uncertainty Modeling):}} Predicting per-pixel uncertainty acts as an adaptive gradient modulator to attenuate corrupted gradients. This successfully removes occlusion artifacts, but without structural priors, geometry in unobserved regions remains noisy and ill-defined.
\emph{\textbf{(3) Exp. C (+ $\mathcal{L}_{spa}$):}} Adding Confidence-Guided Spatial Consistency selectively propagates geometric validity from confident neighbors, restoring a coherent body shape.
\emph{\textbf{(4) Exp. D (+ $\mathcal{L}_{temp}$):}} Incorporating Uncertainty-Weighted Temporal Inertia smooths trajectories and refines surface details, achieving optimal physical plausibility.

\begin{figure}[t]
    \centering
    \includegraphics[width=0.9\linewidth]{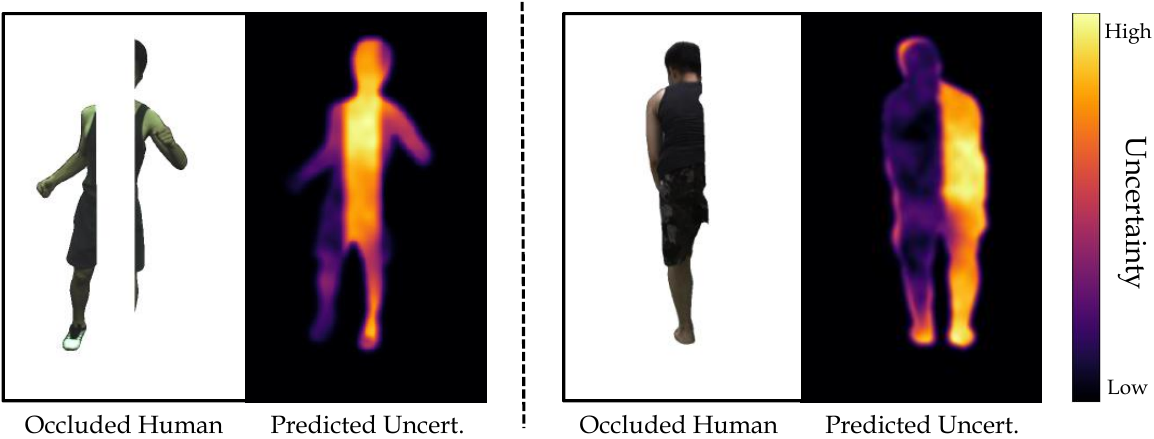} 
    \caption{\textbf{Visualization of the learned Uncertainty Map.} 
    }
    \label{fig:uncertainty_vis}
\end{figure}

\noindent\textbf{Visualization of Uncertainty Maps.}
Fig.~\ref{fig:uncertainty_vis} visualizes the rendered uncertainty map $\hat{U}$ to validate our heteroscedastic noise modeling. As the estimated scale of observation noise, $\hat{U}$ successfully distinguishes valid signals from artifacts. The network assigns low uncertainty to visible body parts, permitting photometric supervision to drive optimization. Conversely, occluded anatomical regions receive high uncertainty. This empirically confirms $\hat{U}$'s role as an adaptive gradient modulator: it increases the NLL denominator (Eq.~\ref{eq:nll}) to attenuate gradients from unreliable cues, effectively preventing canonical geometry from adhering to occluders.

\section{6. Conclusion}
\label{sec:conclusion}

In this paper, we present U-4DGS, a robust framework that reformulates monocular human rendering as an MAP estimation problem to handle severe environmental occlusion. Unlike methods relying on stochastic hallucinations or rigid priors, we explicitly model heteroscedastic observation noise. Our Probabilistic Deformation Network and Joint Rasterization pipeline establish an adaptive gradient modulator, effectively distinguishing valid supervision from occlusion artifacts. Furthermore, Confidence-Aware Regularizations leverage the learned uncertainty to prevent geometric drift in unobserved regions. Extensive evaluations on ZJU-MoCap and OcMotion demonstrate that U-4DGS achieves state-of-the-art robustness and photorealistic fidelity.


\bibliography{aaai2027}


\end{document}